# Bidirectional Attentional Encoder-Decoder Model and Bidirectional Beam Search for Abstractive Summarization

Kamal Al-Sabahi[1], Zhang Zuping[1*], Yang Kang[1]

1.School of Information Science and Engineering, Central South University, China

k.alsabahi@csu.edu.cn, zpzhang@csu.edu.cn, yk_ahead@csu.edu.cn

**Abstract**

Sequence generative models with RNN variants, such as LSTM, GRU, show promising performance on abstractive document summarization. However, they still have some issues that limit their performance, especially while dealing with long sequences. One of the issues is that, to the best of our knowledge, all current models employ a unidirectional decoder, which reasons only about the past and still limited to retain future context while giving a prediction. This makes these models suffer on their own by generating unbalanced outputs. Moreover, unidirectional attention-based document summarization can only capture partial aspects of attentional regularities due to the inherited challenges in document summarization. To this end, we propose an end-to-end trainable bidirectional RNN model to tackle the aforementioned issues. The model has a bidirectional encoder-decoder architecture; in which the encoder and the decoder are bidirectional LSTMs. The forward decoder is initialized with the last hidden state of the backward encoder while the backward decoder is initialized with the last hidden state of the forward encoder. In addition, a bidirectional beam search mechanism is proposed as an approximate inference algorithm for generating the output summaries from the bidirectional model. This enables the model to reason about the past and future and to generate balanced outputs as a result. Experimental results on CNN / Daily Mail dataset show that the proposed model outperforms the current abstractive state-of-the-art models by a considerable margin.

## Introduction

Though our lives have been transformed by ready access to limitless data, we also find ourselves ensnared by information overload (Al-Sabahi et al. 2018, See, Liu, and Manning 2017). It has been said that "It is not information overload, it's filter failure.". To tackle this issue, much attention has been paid to Automatic Document Summarization that alleviates this problem by reducing the size of a long document to a few sentences or paragraphs. Document summarization aims to automatically generate a summary for a document while retaining its original information content.

Recent work in document summarization is dominated by neural network-based methods, namely recurrent neural networks (RNN). Most of which follow the encoder-decoder architecture where a set of recurrent neural network models based on attention encoder-decoder have achieved promising results on short-text summarization (Hou, Hu, and Bei 2018).

**Original Article:** *nicola sturgeon this afternoon scoffed at claims by labour leader ed miliband that he would never do a deal with the snp to become prime minister -* insisting he will ` change his tune ' after the election . the scottish first minister said mr miliband simply ` wo n't have the votes to say that he is going to do what he likes come what may ' and reiterated her call for labour to ` work together to lock the tories out ' . it came after mr miliband this morning repeatedly insisted he would not enter into any deal with the snp after the election - either as a formal coalition or a looser pact to put him in number 1.(…)

**Baseline (See, Liu, and Manning 2017):** scottish first minister said he would never *do a deal with the snp*.it came after mr miliband repeatedly insisted he would not enter into any deal with the snp . but ms sturgeon said ' i suspect ed miliband will change his tune once the votes are cast '

**Bidir_Rev_Cov:** *scottish first minister said mr miliband simply ` wo n't have* the votes to say that he is going to do what may ' and reiterated her call for labour to ` work together to lock the tories out ' it came after mr miliband this morning repeatedly insisted he would not enter into any deal with the snp after the election.

*Figure 1. an example of the generated summaries by the baseline model and ours.*

They proved to be the most efficient methods for the task. RNNs are usually used to play two roles: as an encoder that transforms sequential data into vectors, and as a decoder that transforms the encoded vectors into sequential output (Sun, Lee, and Batra 2017). For a better performance, they also use the attention as an extra input to the decoder. Although encoder-decoder methods can generate sentences with correct grammar, they still have difficulties generating longer sentences with rich semantics. Furthermore, these sentences are also robotic and unable to achieve human's fluency. Since generating longer-text summaries requires higher levels of abstraction while avoiding repetition, it is more challenging, yet more useful (See, Liu, and Manning 2017).

The previous neural text summarization models can be categorized into two common types: extractive and abstractive. Extractive summarization models aim to select and then concatenate a subset of relevant words or sentences that retain the most important information from a document to create its summary (Sun, Lee, and Batra 2017, Al-Sabahi, Zhang, and Nadher 2018). By contrast, abstractive text summarization techniques intend to build an internal semantic representation of the original text and then create a summary closer to a human-generated one, which requires a deeper understanding of the text. Abstractive document summarization is very challenging and still less investigated to date. Since the state-of-the-art abstractive models are still quite weak, most of the previous work has focused on extractive summarization (Al-Sabahi, Zhang, and Nadher 2018, Jeong, Ko, and Seo 2016).

To the best of our knowledge, all the current abstractive models use a unidirectional decoder, usually an RNN variant. When making predictions (in decoding), a unidirectional RNN needs to encode the previous local predictions as a part of the contextual information. If some of the previous predictions are incorrect, the context for the subsequent predictions might include some noise, which undermines the quality of the subsequent predictions. In the example in Figure 1, the baseline model has mistakenly predicted the token "*he*" to come after the sequence "*Scottish First Minister said*". This wrong prediction affects the all the subsequent predictions and leads to wrong information, where *Mr. Miliband* is the one who would never do a deal with the SNP not the *Scottish First Minister, Nicola Sturgeon*. The bidirectionality in our model gives it the ability to handle this kind of issues.

Typically, in unidirectional models, one general way of predicting the next token $y_t$ is to maximize the log probability of $y_t$ given the encoder output and the decoder output generated at the previous time steps, $\log p(y_t|x, Y_{[1:t-1]})$. Since unidirectional models reason only about the past, they are still limited to retain future context $Y_{[t+1:T_y]}$ that can be used for reasoning the previous token $y_t$ by maximizing $\log p(y_t|x, y_{[t+1:T_y]})$. This limitation makes the unidirectional model suffers on its own by not utilizing the past and future dependencies while giving a prediction. In this work, we propose a bidirectional model to remedy these problems. The main contribution of this paper can be summarized as follows:

- We propose a bidirectional encoder-decoder model that has the ability to model both the history textual context and the future one to generate multi-sentence summaries.
- We propose a bidirectional beam search (BBS) as an approximate inference algorithm to decode summaries from the bidirectional model.
- The source sequences are reversed where the output of the forward encoder is fed into to the backward decoder and the output of the backward encoder is fed into the forward decoder.
- Furthermore, we implement the proposed model and we would like to release the source code to the research community.

To the best of our knowledge, we are the first to use bidirectional encoder-decoder topology and bidirectional beam search for the summarization task. Our model differs from the previous models in three important ways: first, both the encoder and the decoder are bidirectional LSTMs. Second, we found that it extremely valuable to reverse the order of the input sequence. Third, a bidirectional approximate inference algorithm (BBS) is used to generate the final output, the summary. Experimental results CNN / Daily Mail dataset have shown that the proposed model outperforms the current state-of-the-art models in terms of ROUGE scores.

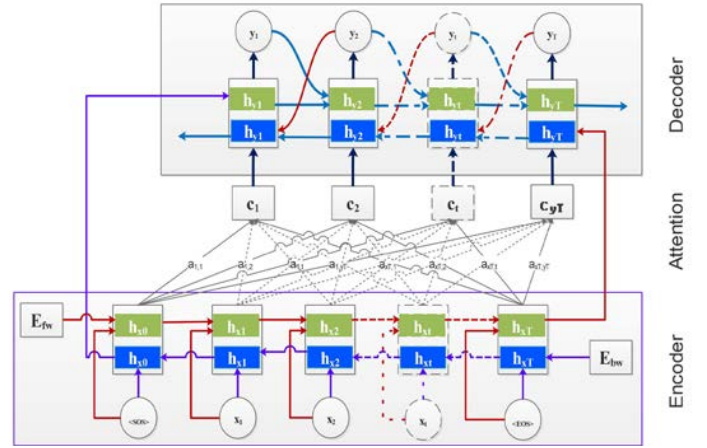

*Figure 2. Bidirectional Encoder-Decoder with attention mechanism. Where $E_{fw}$ and $E_{bw}$ are the initial embeddings from the two directions and $c_t$ is the context vector at the $t$ decoding time step. The encoder outputs are reversed before feeding them to the decoders.*

## The Proposed Model

Recently, generative RNNs are of large interest as they can be used in any NLP task including document summarization (Doetsch, Zeyer, and Ney 2016). Most of which follow the encoder-decoder architecture.

**Definition 1**. Given a document $D$ composed of a sequence of tokens $X = (x_1, x_2, ..., x_{T_x})$ coming from a vocabulary $V$ of size $|V|$, a neural document summarization model is a neural network that directly models the conditional probability $P(Y|X)$ of an optimal sequence $Y = (y_1, ..., y_{T_y})$ that represents a shortened sequence that retain the essence of $X$.

In this section, we describe the proposed model in detail. As shown in Figure 1, there are three main components of our model.

### Bidirectional Encoder-Decoder

In order to consider the context information from the past and future, a bidirectional encoder is used. The input sequence word vectors are fed to LSTM from forward and backward directions.

$$\overrightarrow{h_t^e} = \overrightarrow{LSTM}(x_t, \overrightarrow{h_{t-1}}) \quad (5)$$
$$\overleftarrow{h_t^e} = \overleftarrow{LSTM}(x_t, \overleftarrow{h_{t+1}}) \quad (6)$$

In this work, the decoder is also a bidirectional RNN that composed of two separate LSTM; one decodes the information from left-to-right, forward decoder, while the other decodes from right-to-left, backward decoder. Bidirectionality in RNN on the decoder side supposes to give a better performance. For example, let us say we want to predict the next word "*Roosevelt*" in the sentence "*They said that Teddy Roosevelt was a great President*", on a high level, what a unidirectional LSTM will see is: "*They said that Teddy…*". Trying to predict the next word only by this context is not enough. With bidirectional LSTM, we will be able to see information further down the road, for example:

Forward LSTM: "*They said that Teddy ...*"
Backward LSTM: "*... was a great President.*"

We can see that using the information from the future make it easier for the network to understand that the next word is "*Roosevelt*".

For each decoder, the probability of each target token $(y_1, y_2 ..., y_{T_y})$ is modeled with a SoftMax layer, which transforms the decoder outputs into a probability distribution over a fixed-size vocabulary $V$. This probability is predicted based on the decoder recurrent state and the previously generated token. From the attention layer, a vector called the context vector $c_t$ is given as additional input to the decoder, calculated using Eq. 12.

The objective of the model training is to find the best parameter set, $\theta$, that maximize the conditional probability of the sentence-summary alignment in the training corpus from both directions. Given the previous tokens, the model factorizes the conditional into a summation of individual log conditional probabilities from both directions, Eq. 3.

$$P(y_t|[y_d]_{d \neq t}) = \overrightarrow{\log p(y_t|Y_{[1:t-1]})} + \overleftarrow{\log p(y_t|Y_{[t+1:T_y]})} \quad (3)$$

Where $\overrightarrow{\log p(y_t|Y_{[1:t-1]})}$ and $\overleftarrow{\log p(y_t|Y_{[t+1:T_y]})}$ are the left-to-right and the right-to-left LSTM decoder model, Eq. 4 and Eq. 5 respectively.

$$\overrightarrow{\log p(y_t|Y_{[1:t-1]})} = \sum_{t=1}^{T_y} \log p(y_t|\{y_1, ..., y_{t-1}\}, x; \vec{\theta}) \quad (4)$$
$$\overleftarrow{\log p(y_t|Y_{[t+1:T_y]})} = \sum_{t=1}^{T_y} \log p(y_t|\{y_{t+1}, ..., y_{T_y}\}, x; \overleftarrow{\theta}) \quad (5)$$
$$p(y_t|\{y_1, ..., y_{t-1}\}, x; \vec{\theta}) = g(y_{t-1}, \overrightarrow{h_t^d}, c_t) \quad (6)$$
$$\overrightarrow{h_t^d} = LSTM(y_{t-1}, \overrightarrow{h_{t-1}^d}, c_t) \quad (7)$$
$$p(y_t|\{y_{t+1}, ..., y_{T_y}\}, x; \overleftarrow{\theta}) = g(y_{t+1}, \overleftarrow{h_t^d}, c_t) \quad (8)$$
$$\overleftarrow{h_t^d} = LSTM(y_{t+1}, \overleftarrow{h_{t+1}^d}, c_t) \quad (9)$$

At training time, in the forward decoder, $y_{t-1}$ is the previous token in the reference summary, while at testing time, it is the previously generated token, the previous decoder output. $h_{t-1}$ is the previous hidden state, and $c_t$ is context vector calculated as a weighted average of the encoder hidden states $h_1^e, ..., h_{T_x}^e$ using the attention mechanism, as in Eq. 10, Eq. 11, & Eq. 12. The same applies to the backward decoder.

$$e_{ij} = v^T \tanh(W_h^d h_i^d + W_h^e h_j^e + b_{attn}) \quad (10)$$
$$\alpha_{ij} = \frac{\exp(e_{ij})}{\sum_{k=1}^{T_x} e_{ik}} \quad (11)$$
$$c_i = \sum_{j=1}^{T_x} \alpha_{ij} h_j \quad (12)$$

In Eq. 12, there are two subscripts for $e_{ij}$ scores. The first one is $i$, which represents the output time step, and $j$ represents the input time step.

According to the works of (Doetsch, Zeyer, and Ney 2016, Sutskever, Vinyals, and Le 2014) for Machine Translation problem, it has been discovered that LSTM learns much better when the source sequence is reversed, while the target one is not reversed. We hypothesize that this reverse connectivity will enhance the performance of the LSTM by making it easier for Stochastic Gradient Descent to establish a communication between the input and the output (Sutskever, Vinyals, and Le 2014). In this work, the backward decoder is initialized with the final state of the forward LSTM encoder, and the forward decoder is initialized with the final state of the backward LSTM encoder. We observed an improvement while applying this to the proposed model.

## Handling Out-of-Vocabulary and repetition

To handle the Out-of-Vocabulary (OOV) problem, we borrowed the pointer-generator network from (See, Liu, and Manning 2017) which has the ability to smartly choose between copying words from the source text or generating novel ones. Furthermore, we adopted the coverage mechanism from (See, Liu, and Manning 2017). We refer the reader to the original article for more detail.

## The Objective Function

The proposed model is trained using the stochastic gradient descent (SGD). We use the following joint loss function:

$$\ell = \gamma \ell^f + (1-\gamma) \ell^b \quad (13)$$

where $\gamma$ is a scaling factor used to define the difference in magnitude between the forward and backward losses. We have tried different range of values in the interval [0,1]. The best value was $(\gamma = 0.7)$. $\ell^f$ and $\ell^b$ are the forward and backward losses computed by accumulating the SoftMax losses of the forward and backward directions respectively. The objective is to minimize the joint loss $\ell$, which means, in other words, maximizing the probability distribution of the correct summary by finding an optimal sequence that maximizes the following scoring functions, Eq. 14 & Eq. 15:

$$F^f(\vec{\theta}) = \frac{1}{T_y} \sum_{t=1}^{T_y} \log p(y_t|Y_{[1:t-1]}, x; \vec{\theta}) \quad (14)$$
$$F^b(\overleftarrow{\theta}) = \frac{1}{T_y} \sum_{t=1}^{T_y} \log p(y_t|Y_{[t+1:T_y]}, x; \overleftarrow{\theta}) \quad (15)$$

Where $Y_{[1:t-1]}$ and $Y_{[t+1:T_y]}$ are the previously generated words from the forward and backward decoders respectively, and $x$ is the input sequence.

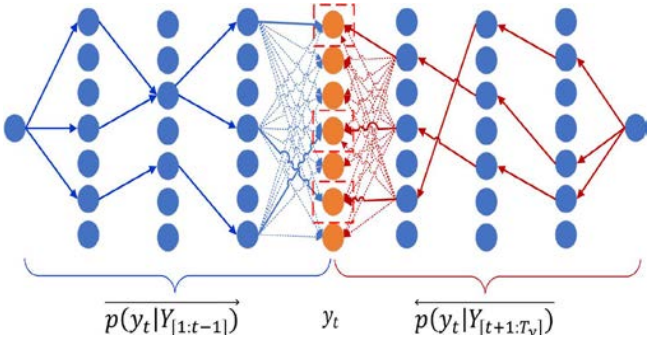

Figure 3. Bidirectional beam search. In which the decision on $y_t$ conditioned on the tokens from left-to-right and the right-to-left directions.

**Bidirectional Beam Search**

One of the concerns about the previous beam-search decoding models is that they consider only the information from the past. Relying on information from the future seems at first sight to violate causality. How can we base our understanding of what we have read on something that has not been said yet? However, human readers do exactly that. Words and even whole sentences, which at first mean nothing, are found to make sense in the light of the future context. The use of bidirectional LSTMs can offer some benefit in terms of better results to sequence prediction problems, where it is appropriate.

In this work, we want the Bidirectional Beam Search (BBS) to reason about both forward and backward time dependencies. We consider this under the hypothesis that the performance of the model on generating the summary will benefit from combining the information from the past and future. This mimics what the human does when summarizing a document. Where he/she read forward and backward looking for the best information that better represent the document under the limitation of the predefined summary length (compression rate).

The toughest challenge for applying Bidirectional decoder is the fact that we should have to know the complete sequence to get it working. A thing that is difficult in the testing phase. To overcome this issue, we run a unidirectional backward beam search to get the initial full backward sequence. Then apply the forward beam search considering the information from the backward direction, as shown in Fig. 2 and Eq 17. The algorithm of the proposed beam search, BBS, works as follows:

(1) We run the backward beam search to get the full backward generated sequence $\overleftarrow{B}_{[1:T_y]}$, Eq. 16.

(2) We fed this sequence to the forward beam search. At the beginning, we started with start decoding token and the full backward beams. Then the forward steps will increase while the backward steps will decrease. As in Eq. 17, the forward beams at $t$ time step is computed as the summation of the forward beams and backward beams computed at step 1. In the forward beam search, making a prediction of $y_t$ is a joint search approximation, so it will not depend only on the information from the past but also on the information from the future. The new search space at time step $t$ is $V_t = (B_{[1:t-1]} \times V) + (V \times B_{[t+1:T]})$.

(3) The output of the joint search is sorted and the best one is picked as output.

(4) The log probability of a token is normalized by the number of tokens, otherwise the longer sequences will always have lower probabilities.

$$\overleftarrow{B_{[1:T_y]}} = topk \sum_{i=1}^{T_y} \log p(y_i|Y_{[i+1:T_y]}) \quad (16)$$

$$Y_{[1:T_y]} = \sum_{t=1}^{T_y}(\gamma \sum_{i=1}^{t} \log p(y_i|Y_{[1:i-1]}) + (1-\gamma) \sum_{j=t}^{T_y} \log p(y_j|\overleftarrow{B_{[j+1:T_y]}})) \quad (17)$$

Where $\gamma$ is a scaling values defining the magnificent of the forward and backward beams. By experiment, the best value is $\gamma = 0.7$.

**Remark 1.** To avoid notation clutter in Eq. 16 & Eq. 17, we assume that the beam search size is equal to 1, ($k = 1$) and the length of the forward and backward beams are equal to $T_y$. Though, this is not the case in our experiment where the beam size was set to 4 and since the algorithm is self-stop, the length of the beams may vary.

The new beam search is a new way of approximating the solution considering the context from the past and future while predicting the output sequence. Bidirectional Beam Search was firstly proposed by (Sun, Lee, and Batra 2017) for the task of fill-in-the-blank image captioning. The main difference between our work and the one of (Sun, Lee, and Batra 2017) comes in threefold: the first is the fact that they used both forward and backward for initialization and then they alternated between the forward and backward beam searches with respect to the approximate joint search that we think is not necessary where the forward initialization will be overridden, in this context, during the joint beam search. The second is the complexity of our model is lower. As we will discuss later, the time complexity of our BBS is $3|B||V|$ while the one of (Sun, Lee, and Batra 2017) is $|B||V||B|$. The third difference lies on the fact that they applied their model on filling-the-blank for image captioning task, which is much simpler than our problem, document summarization, in the fact that there is more information in the input of the decoder and in most cases, they just need to find the best single token in the vocabulary to fill the gap. In contrast, document summarization is much complex and challenging since we need to generate the whole sequence from scratch.

### The complexity Analysis

The complexity of the proposed method comes in twofold: the first one related to the bidirectional decoder that adds extra parameters. For a vocabulary size of 50k, the baseline model (See, Liu, and Manning 2017) has 21,501,265. The proposed model has 34,434,722 parameters. The second one related to the bidirectional beam search. As we mentioned earlier, the proposed BBS has two main processes. The first one is backward beam search, which is a standard beam search with a time complexity of $|B||V|$. The second is the bidirectional beam search, which is a combination of forward and backward beams calculated at a single round of BBS with a time complexity of $|B||V| + |V||B|$ comes from the forward and backward beam searches respectively. The overall time complexity is $|B||V| + |B||V| + |V||B|$ which roughly equal $3|B||V|$. It is worth mentioning that the difference is trivial between the complexity of the proposed BBS and the standard beam search.

## Experiments

### Datasets

In this work, we used CNN/Daily Mail dataset to evaluate the proposed model. CNN/Daily Mail dataset was originally built by (Hermann et al. 2015) for question answering task and then re-used for extractive (Cheng and Lapata 2016, Nallapati, Zhai, and Zhou 2017, Al-Sabahi, Zhang, and Nadher 2018) and abstractive text summarization tasks (Nallapati et al. 2016, Paulus, Xiong, and Socher 2017, See, Liu, and Manning 2017). In the joint CNN/Daily Mail dataset, there are 286,726 for training, 13,368 for validation and 11,490 for testing. The source documents in the training set have 781 words spanning 29.74 sentences on an average while the summaries consist of 56 words and 3.75 sentences. We follow the work of (See, Liu, and Manning 2017) to obtain the same version of the dataset. The documents in this dataset are considered as long documents, and the summaries are multi-sentence ones that bring challenges in building powerful models that generate novel and readable summaries.

### Baselines

Several abstractive and extractive based summarizations have been proposed recently. We choose the ones that are comparable to our work. Six common baselines are used; some of which are the current state-of-the-art.

- **Lead-3** baseline: it simply chooses the first three sentences of the document as a summary. Since we are working on news article datasets, it is usual for the important information to be put at the beginning of the article. With this have been said, the LEAD-3 model tends to have good ROUGE scores in news-based datasets.
- **SummaRuNNer** (Nallapati, Zhai, and Zhou 2017) is used as extractive baselines.
- **RL**: a reinforced abstractive summarization model (Paulus, Xiong, and Socher 2017), a **Pointer-Generator** based Network (See, Liu, and Manning 2017), **SummaRuNNer-abs** (Nallapati, Zhai, and Zhou 2017), and **words-lvt2k** (Nallapati et al. 2016) are used in this work as abstractive baselines.

### Experiment Settings

The word embedding is learned from scratch on the joint CNN/DailyMail dataset with a dimension of 128. The model hidden state size is set to 256. The vocabulary size is limited to 50k for both the source and target. We use Adagrad (Duchi, Hazan, and Singer 2011) to train the model with a batch size of 32, a learning rate of 0.15 and initial accumulator value of 0.1. No form of regularization has been used, but the gradient clipping with a maximum gradient norm of 2. We implement the early stopping based on the loss on the validation set. The articles are truncated to 400 tokens and the length of the summary is limited to 100 tokens in the training and testing. The model is trained on a single NVIDIA TITAN Xp 12 GB GPU. At test time, we used the proposed BBS with a beam size of 4 to generate the summaries. We save model checkpoints every 6 minutes and choose the best checkpoint based on the running average loss on the validation set. Following the work of (See, Liu, and Manning 2017), we train the models with coverage as a separate training phase. We stopped training after the coverage loss converge down to 0.17 from the initial value (0.5). Table 3 shows the number of iterations and the training time for each model variant.

### Experimental Results

ROUGE Toolkit (Lin 2004) and Pyrouge python package are used to evaluate the performance of the proposed models. ROUGE-1, ROUGE-2, and ROUGE-L scores are reported. The proposed model variants, as shown in Table 1, are compared with several abstractive and extractive baselines, mentioned in the baseline section. The evaluation results in Table 2 show that the proposed model achieved promising results. From which, we can make the following observations:

- The first two models in Table 2 are the extractive baselines while the rest are abstractive ones and the proposed models are in the lower part. In ROUGE setting, we used -c 95 option, which means that all our ROUGE scores have 95% confidence interval of at most ±0.25. The models marked with (*) are evaluated on the anonymous dataset, which means they are not literally comparable to the proposed models.

- We get the values in Table 3 by concurrently running the training on the train set and the evaluation on the validation set and saving the best models on the validation set so far for the later early stopping implemention.
- The results in Table 2 show that LEAD-3 model achieved competitive scores. A possible reason is in a news article, it is usual for the important information to be put at the beginning of the news article. This makes this baseline hard to be beaten; however, our model has performed better.

| Bidir_NoRev_No-Cov | Bidirectional Enc-Dec + Non-Reversed input sequence |
|---|---|
| Bidir_Rev_NoCov | Bidirectional Enc-Dec + Reversed input sequence |
| Bidir_NoRev_Cov | Bidirectional Enc-Dec + Non-Reversed input sequence+Coverage |
| Bidir_Rev_Cov | Bidirectional Enc-Dec + Reversed input sequence+Coverage |

*Table 1. The proposed model variants*

- The values in Table 2 and Table 3 show that the models with reversed input sequence, Bidir_Rev_NoCov and Bidir_Rev_Cov, achieve better ROUGE scores and require less iterations compared to their counterparts with the original sequence. This assert that LSTM learns much better when we feed the output of the forward encoder as an input into the backward decoder and the output of the backward encoder into the forward decoder, which proves our earlier hypothesis. The most probable reason for this is that in news articles, the writers tend to deliver the most important information at the beginning of their piece of writing. Reversing the source sequence helps the model to make more confident predictions in the early time steps and less confident predictions in the later parts. This will help much with the longer sentences where reversing the sequence aid the LSTM memory utilization.

| Models | R-1 | R-2 | R-L |
|---|---|---|---|
| LEAD-3 (See, Liu, and Manning 2017) | 40.3% | 17.7% | 36.6% |
| SummaRuNNer (Nallapati, Zhai, and Zhou 2017)* | 39.6% | 16.2% | 35.3% |
| SummaRuNNer-abs (Nallapati, Zhai, and Zhou 2017)* | 37.5% | 14.5% | 33.4% |
| words-lvt2k (Nallapati et al. 2016)* | 32.5% | 11.8% | 29.5% |
| Pointer-Gen (See, Liu, and Manning 2017) | 36.4% | 15.7% | 33.4% |
| Pointer-Gen+Coverage (See, Liu, and Manning 2017) | 39.5% | 17.3% | 36.4% |
| RL, with intra-attention (Paulus, Xiong, and Socher 2017) | 41.6% | 15.7% | **39.1%** |
| Ours (Bidir_NoRev_NoCov) | 37.2% | 16.1% | 33.5% |
| Ours (Bidir_Rev_NoCov) | 39.1% | 16.4% | 35.1% |
| Ours (Bidir_NoRev_Cov) | 40.1% | 17.9% | 37.2% |
| Ours (Bidir_Rev_Cov) | **42.6%** | **18.8%** | 38.5% |

*Table 2. The full-length F1 ROUGE scores on CNN/Daily Mail test set with respect to the baselines*

- The models that do not use the coverage mechanism, Bidir_NoRev_NoCov, Bidir_Rev_NoCov, suffer from the repetition problem. While the models with coverage, Bidir_NoRev_Cov, Bidir_Rev_Cov, achieve better scores and the repetition problem has been adequately addressed.
- ROUGE metrics measure the n-gram overlap between a reference summary, usually generated manually by a human, and a system generated one. This nature of ROUGE metrics makes it possible for the generative models that optimizes for ROUGE metric to obtain high ROUGE scores compromising on the quality and the readability of the generated summaries (Paulus, Xiong, and Socher 2017, Liu, Lowe, et al. 2016).

| Models | Days | Hours | #Iterations | #Epochs |
|---|---|---|---|---|
| Bidir_NoRev_NoCov | 3 | 7 | 136,833 | 15.3 |
| Bidir_Rev_NoCov | 2 | 21 | 118,857 | 13.3 |
| Bidir_NoRev_Cov | 3 | 16 | 151,394 | 16.8 |
| Bidir_Rev_Cov | 3 | 6 | 134,310 | 14.9 |

*Table 3. The training time and the number of iterations for the best models for each variant of the proposed model*

- To sum up, the ROUGE-1, ROUGE-2, and ROUGE-L scores, in Table 2, assert that the proposed model, Bidir_Rev_Cov, achieved the best on CNN/Daily Mail outperforming the other model variants and the baselines. This indicates that the combination of a bidirectional encoder-decoder topology, reversed input sequence, self-attention and a bidirectional beam search leads to a better performance on the abstractive summarization task and yield state-of-the-art performance.

## Qualitative Analysis

In addition to ROUGE Scores, we chose some summaries generated by different model variants. In the favor of the limited space, we include only one example in Figure 4, which show that the proposed model has the ability to generate balanced output. In the context of summarization, the unbalanced output means that the generated summary captures only part of the essential information while neglecting the other parts. Since we are limited by the compression rate, the maximum decoding steps, the decoding process might stop while the important information has yet to come. Letting the forward and backward decoding process vote on the predicted token is supposed to tackle this issue. A closer look at the generated summaries in Figure 4 asserts that the baseline model couldn't capture the whole context while the proposed model handles this issue seamlessly.

## Related Work

Deep learning methods are widely used in natural language processing tasks, such as sentiment analysis, machine translation, question answering, textual entailment, and document summarization. In this section, we present a brief description for some of the previous works that are comparable to ours.

**Reference Summary** erica avery , 17 , was cuffed and hauled out of a broward county , florida courtroom after being accused of using the internet on tuesday. she was taken into custody barely a month after making $ 100,000 bond on charges she aided in the rape of a friend in 2013.she and four other teens are charged with armed sexual battery and kidnapping in connection with the assault on a victim in hollywood .

**Baseline: (See, Liu, and Manning 2017) (ROUGE-1 0.453)** erica avery was cuffed and hauled out of a broward county courtroom after being accused of using the internet , which itself violated the conditions set just one month prior when she was first released on $ 100,000 bond. prosecutor maria schneider said photos of the 17-year-old were posted to social media and one of them shows her using a smart phone , reports the sun-sentinel.the barbaric attack was filmed on a cell phone by lanel singleton .

**Bidir_NoRev_NoCov (ROUGE-1 0.397):** erica avery hauled out of a broward county courtroom after being accused of using the internet , which itself violated the conditions set just one month prior when she was first released on $ 100,000 bond .she was accused of using the internet , which itself violated the conditions set just one month prior when she was first released on $ 100,000 bond .prosecutor maria schneider said photos of the 17-year-old were posted to social media and one of them shows her using a smart phone , reports the sun-sentinel .

**Bidir_Rev_NoCov (ROUGE-1 0.467)** erica avery was cuffed and hauled out of a broward county courtroom after being accused of using the internet, which itself violates the conditions set just one month prior when she was first released on $ 100,000 bond. conditions set just one month prior when she was first released on $ 100,000 bond .

**Bidir_NoRev_Cov (ROUGE-1 0.523)** one of several florida teens accused of ganging up on a friend by allowing her to be sexually assaulted was thrown back in jail tuesday for allegedly breaking the conditions of her pre-trial release by threatening the victim online.erica avery was cuffed and hauled out of a broward county courtroom after being accused of using the internet, which itself violated the conditions set just one month prior when she was first released on $ 100,000 bond .

**Bidir_Rev_Cov: (ROUGE-1 0.770)** erica avery was cuffed and hauled out of a broward county courtroom tuesday after being accused of using the internet , which itself violates the conditions set just one month prior when she was first released on $ 100,000 bond.avery and four other teens are charged with armed sexual battery and kidnapping in connection with the 2013 assault on a victim in hollywood , florida .

*Figure 4. Example of abstractive summaries generated by the proposed model variants.*

## Bidirectional RNN

Generative models with encoder-decoder architecture have recently attracted an extensive interest. In it is basic form, the input is encoded into a vector representation. The encoder output is used as an initial input to the decoder. An RNN variant, such as LSTM, is used as a decoder that sequentially generates the output. Most of the previously proposed encoder-decoder models used a bidirectional encoder and unidirectional decoder. The bidirectional decoder topology was used in a few works to solve some specific tasks. For example, a work proposed by (Wang et al. 2016) used a bidirectional decoder as a try to make use of the past and future context information during generating an image caption. Two embeddings were used, sentence embeddings and visual embeddings. The sentence embeddings were encoded using a bidirectional LSTM and the visual embeddings were encoded by CNN. Furthermore, they used a deep bidirectional LSTM architecture to learn higher level embeddings. A standard beam search algorithm was used to select the optimal output sequence. Another work has been proposed by (Liu, Finch, et al. 2016), in which they utilized the agreement between a pair of target-directional LSTMs, one generates sequences from the left-to-right and the other generates sequences from right-to-left, to generate more balanced outputs. Moreover, they introduced two approximate search models which used only a small subset of the entire search space. Their paper was addressing the machine transliteration and grapheme-phoneme problems. Qing Sun (Sun, Lee, and Batra 2017) presented Bidirectional Beam Search to address the Fill-in-the-Blank Image Captioning task. They used both past and future sentence structure to recreate sensible picture depictions. They started by decomposing the bidirectional RNN into two unidirectional RNN. Then, a beam search was performed on one direction while holding the beams in the other direction fixed. We discussed the difference between this work and ours earlier.

## Document Summarization

The recent advance in neural network architecture and training algorithms spark a huge interest in applying the deep learning models to solve abstractive summarization problem. Several neural network-based approaches have been proposed for abstractive document summarization. Most of which, such as (Rush, Chopra, and Weston 2015, Nallapati et al. 2016, Paulus, Xiong, and Socher 2017, See, Liu, and Manning 2017), followed the common encoder-decoder architecture. They differ in the terms of which RNN variant was used for the decoder and how the encoder encoded the source document representation.

The first work was carried out by (Rush, Chopra, and Weston 2015) in which they proposed an encoder-decoder model where the encoder was a convolutional network and the decoder was a feedforward language model. The convolutional encoder was enhanced by an attention mechanism. The trained model was used, then, as a feature to a log-linear model. One of the shortcomings of this approach is the fact that just the first sentence of each news article was used to generate its title. Paulus et al (Paulus, Xiong, and Socher 2017) have introduced an abstractive model with a new training objective combining the maximum-likelihood cross-entropy with a reward from policy reinforcement learning (RL). This approach optimizes for a single discrete evaluation metric, ROUGE, which may lead to a good ROUGE score but poor output summary. Another interesting work was proposed by (See, Liu, and Manning 2017). They used a hybrid pointer-generator network to copy the words from the input document to improve the accuracy and to tackle OOV problem. Moreover, they introduced a mechanism for the coverage vector that alleviated the repetition problem in the generated output. The model followed the standard encoder-decoder architecture.

The performance of RNN-based encoder-decoder models is quite good for short input and output sequences; however, for longer documents and summaries, these models often struggle with serious problems such as repetition, unreadability, and incoherence. Moreover, they tend to generate unbalanced output. As we mentioned earlier in the paragraph before the last one in the introduction section, our proposed model is different from the related work in the sense that it utilizes both the long-term history and future context by using the bidirectional encoder-decoder architecture. Moreover, it uses a bidirectional beam search that infers the generated summary from the past and future. This gives the model the capability to reason about the past and future.

The closest model to ours is the one proposed by (See, Liu, and Manning 2017). From which we borrowed the pointer-generator network and the coverage models to handle the OOV and repetition respectively. Different from that of (See, Liu, and Manning 2017) the decoder and the beam search of our model are bidirectional. Furthermore, unlike the one of (Paulus, Xiong, and Socher 2017), our model neither use any attention in the decoder part nor any kind of reinforcement learning that optimize to the discrete values of ROUGE scores.

## Conclusion and Future Work

In this work, we used a bidirectional encoder-decoder architecture; each of which is a bidirectional recurrent neural network consists of two recurrent layers, one for learning history textual context and the other for learning future textual context. The output of the forward encoder was fed as input into the backward decoder while the output of the backward encoder was fed into the forward decoder. Then, a bidirectional beam search mechanism is used to generate tokens for the final summary one at a time. The experimental results have shown the effectiveness and the superiority of the proposed model compared to the-state-of-the-art models. Even though the pointer-generator network has alleviated the OOV problem, finding a way to tackle the problem while encouraging the model to generate summaries with more novelty and high level of abstraction is an exciting research problem. Furthermore, we believe that there is a real need to propose an evaluation metric besides ROUGE to optimize on summarization models, especially for long sequences.

9